# CLIP Model for Images to Textual Prompts Based on Top-k Neighbors

Xin Zhang[1],
University of California,Santa Cruz,
Santa Cruz, United States,
xzhan445@ucsc.edu,

YeMing Cai[2],
Wuhan, China,
Wuhan University, China,
2018282110317@whu.edu.cn,

Xin Zhang[1],
University of California,Santa Cruz,
Santa Cruz, United States,
xzhan445@ucsc.edu,

Tianzhi Jia[*],
Beijing Jiaotong University,
Beijing, China,
iatianzhi@bjtu.edu.cn

**Abstract—** Text-to-image synthesis, a subfield of multimodal generation, has gained significant attention in recent years. We propose a cost-effective approach for image-to-prompt generation that leverages generative models to generate textual prompts without the need for large amounts of annotated data. We divide our method into two stages: online stage and offline stage. We use a combination of the CLIP model and K-nearest neighbors (KNN) algorithm. The proposed system consists of two main parts: an offline task and an online task. Our method owns the highest metric 0.612 among these models, which is 0.013, 0.055, 0.011 higher than Clip, Clip + KNN（top 10） respectively.

## I. Introduction

Text-to-image method, a subfield of multimodal generation, has gained significant attention in recent years. [1,2,3] This emerging field focuses on generating realistic and visually coherent images from textual descriptions. The advancements in deep learning and generative models have propelled the popularity of text-to-image synthesis, enabling the creation of high-quality and diverse visual content.

Among many Text-to-image methods, Stable Diffusion is a significant open-source model. The core architecture of Stable Diffusion consists of three main models: the autoencoder, the CLIP text encoder, and the U-Net. These models work in tandem to facilitate the generation of high-quality and diverse samples. However, there are some limitations in textual prompts. The main challenge lies in the fact that the details and features in the reference image often cannot be fully captured through simple textual descriptions. These details may involve aspects such as color, texture, shape, and position, which are difficult to accurately convey through text prompts alone. Moreover, there may be limitations in people's linguistic expression, preventing them from precisely describing the complex visual information present in the reference image.

We propose a cost-effective approach for image-to-prompt generation that leverages generative models to generate textual prompts without the need for large amounts of annotated data. Our method allows for direct utilization of the generated prompts or serves as valuable initialization for data-efficient fine-tuning processes. This approach significantly reduces data costs and time consumption while achieving high quality and diversity in the generation of prompts related to input images.

In this paper, we utilize CLIP and KNN for prompt prediction. Related work is described in section II, and we introduce our methodology and experiment in section III and IV.

## II. Related Work

Stable diffusion [4] is a latent text-to-image diffusion model. In the diffusion model, the initial input is a random Gaussian noise, and through a series of iterative steps, the model gradually reduces the impact of noise through the learning and optimization process, generating outputs that are increasingly similar to the target sample. In each diffusion step, the model introduces some randomness and control parameters to ensure that the output progressively approaches the target sample while maintaining the coherence and details of the image.

However, the diffusion model has a high computational complexity, requiring a large number of iterative steps, which leads to significant time and computational resource consumption during sample generation. The latent diffusion model improves upon the traditional diffusion model by applying the diffusion process in a lower-dimensional latent space, thereby reducing the storage and computational complexity. In contrast to the traditional diffusion model that operates directly in the pixel space, the latent diffusion model generates compressed latent representations, effectively reducing the dimensionality of the data.

There are three main components in latent diffusion:

an auto encoder (VAE)[5], a U-Net[6], CLIP[7]'s Text Encoder. The whole structure is shown in Fig 1.

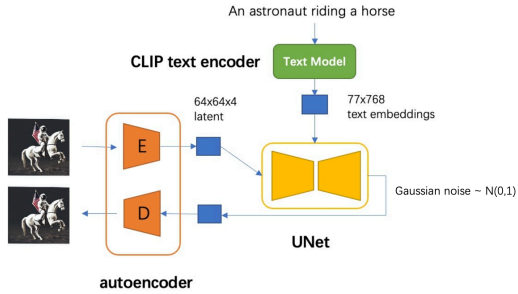

Fig 1: Stable Diffusion framework

The VAE model comprises an encoder and a decoder component. The encoder is responsible for mapping the input image to a low-dimensional latent representation, which serves as the input to the U-Net model. Conversely, the decoder reconstructs the image from the latent representation.

The U-Net model consists of an encoder and a decoder, both composed of ResNet blocks. The encoder compresses the image representation into a lower-resolution image representation, while the decoder decodes the low-resolution image representation back to the original high-resolution image representation, aiming to reduce noise. Specifically, the output of the U-Net is used to predict the noise residual, which is then used to compute the predicted denoised image representation. To prevent the loss of important information during down sampling, shortcut connections are typically added between the down sampling ResNet blocks in the encoder and the up-sampling ResNet blocks in the decoder. Furthermore, the stable diffusion U-Net is capable of incorporating text embeddings through cross-attention layers. These cross-attention layers are typically added between the ResNet blocks in both the encoder and decoder parts of the U-Net.

The text encoder plays a crucial role in converting input prompts into a latent embedding space that is interpretable by the U-Net model. Typically, the text encoder adopts a simple yet powerful Transformer-based architecture to process the input token sequence and generate a series of latent text embeddings. The CLIP Text encoder, commonly employed in this context, leverages its pre-trained knowledge and language understanding capabilities to capture the semantic information and contextual relationships within the text prompts. By effectively encoding the textual information into a latent representation, the U-Net model can establish a meaningful correspondence between text and image modalities, facilitating the generation of visually coherent and contextually aligned outputs.

- Our Contribution
- ✓ We hybrid LightGBM and XGBoost to predict the credits' Default.
- ✓ We introduce our dataset and do some analysis.
- ✓ In the experiments process, we do the comparing experiments and the result shows that our model performed better than the other models.

III. METHODOLOGY

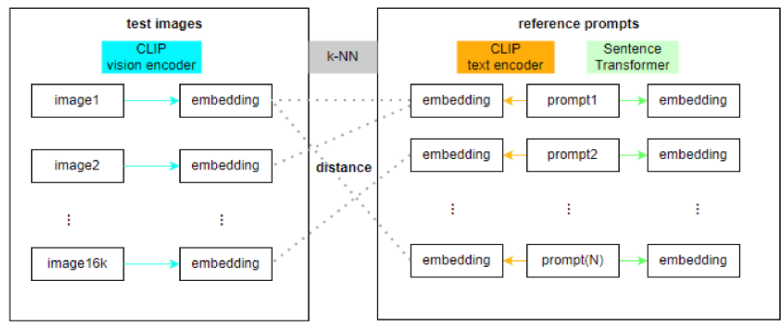

We propose a method that transfer images to textual prompts based on top-k neighbors. Our method allows for direct utilization of the generated prompts or serves as valuable initialization for data-efficient fine-tuning processes. It can be seen as two stages: offline stage and online stage.

- Offline stage

At the offline stage, firstly, we collect a dataset $D_{\{prompt\}}$ consisting of prompt texts, where each prompt text is denoted as $\{prompt_i, i = 1, 2, \ldots, N\}$. These prompt texts are gathered from diverse and reliable sources, including books, news articles, online forums, among others. We ensure that the collected data covers various domains and topics to improve the model's generalization ability.

Next, to transform the prompt texts into interpretable embedding representations, we utilize the text encoder component of the CLIP model. This encoder converts the input prompt text into corresponding CLIP text embeddings $E_{CLIP_{text}}(prompt_i)$. These embedding vectors capture the semantic information and contextual associations of the text, enabling the model to better understand and process the prompt texts.

Additionally, to further enhance the model's understanding of the text, we employ the Sentence Transformer model for sentence embeddings. This process converts each prompt text $prompt_i$ into sentence text embeddings $E_{sent}(prompt_i)$, leveraging deep learning models that capture the semantic and contextual information of sentences.

By collecting and transforming these prompt texts, we construct a dataset $D_{pre}$ containing prompt texts $\{prompt_i, i = 1, 2, \ldots, N\}$, CLIP text embeddings $\{E_{CLIP_{text}}(prompt_i), i = 1, 2, \ldots, N\}$, and sentence text embeddings $E_{sent}(prompt_i), i = 1, 2, \ldots, N$. This dataset provides rich reference information for subsequent online tasks, allowing the model to accurately match and calculate similarity between input images and text. By leveraging these embedding vectors, our model can better comprehend the input prompts and generate accurate and meaningful output results.

- Online stage

In the online stage, we aim to generate corresponding textual prompts for the input images. This process involves the following steps:

Firstly, we employ the image encoder component of CLIP to encode the input images, resulting in the image embedding representation $\{E_{CLIP_{image}}(image_i), i = 1, 2, \ldots, N\}$. This embedding vector captures the semantic and feature information of the image.

To find the most similar text to the input image, we employ the K-Nearest Neighbors (KNN) algorithm to search the stored CLIP Text Embedding database $\{E_{CLIP_{text}}(prompt_i), i = 1, 2, \ldots, N\}$ for the top K CLIP Text Embeddings most similar to the input image embedding. These top K embedding vectors are denoted as $\{E_{CLIP_{text}}(prompt_j), j = 1, 2, \ldots, k\}$.

Subsequently, we compute the average of these top K embedding vectors to obtain the predicted embedding $E_{pred1}$, as shown below:

$$E_{\text{pred1}} = \frac{1}{K} \sum_{j=1}^{K} E_{CLIP_{text}}(prompt_j)$$

In addition, we use the CLIP model to get the prompt $prompt_m$ corresponding to the image, and utilize the Sentence Transformer model, such as all-LM-V6 to compute the prompts $E_{pred2}$ with CLIP.

$$E_{\text{pred2}} = E_{sent}(prompt_m)$$

Then we combine $E_{pred1}$ and $E_{pred2}$ to get $E_{pred}$.

$$E_{\text{pred}} = w_1 * E_{\text{pred1}} + w_2 * E_{\text{pred2}}$$

## IV. Experiments

- Experiments data

We utilize some existing image-prompts datasets: latin400m [8], coyo700[9], COCO[10], and red caps[11].

*1)* Latin400m: dataset containing latin texts for machine learning, language generation and analysation.

*2)* COYO700: COYO-700M is a large-scale dataset that contains 747M image-text pairs as well as many other meta-attributes to increase the usability to train various models.

*3)* COCO: The MS COCO (Microsoft Common Objects in Context) dataset is a large-scale object detection, segmentation, key-point detection, and captioning dataset.

*4)* RedCaps: RedCaps is a large-scale dataset of 12M image-text pairs collected from Reddit.

- Feature engineering

After selecting the textual data, a rigorous cleaning process was applied to remove non-English text descriptions, empty texts, texts containing NaN or other exceptional vocabulary. This process resulted in approximately 3.2 million remaining textual data. Subsequently, the SDv2 model was employed to generate corresponding images based on the cleaned dataset. Following the generation of the dataset, a secondary data cleaning step was conducted, filtering out prompts with a cosine similarity above 0.9 (calculated based on corresponding embeddings). Only a subset of prompts was retained. Additionally, statistical analysis was performed on the prompt vocabulary, discarding prompt terms with a similarity below 0.6 compared to the vocabulary of the all-MiniLM-L6-v2 model. This alignment with the testing set ensured further quality assurance of the embeddings.

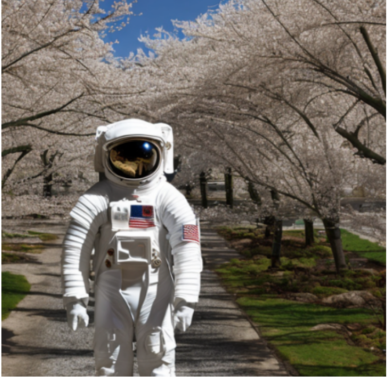

an astronaut standing on a engaging white rose, in the midst of by ivory cherry blossoms

Figure 3: variables relationship

- Training parameters

The model's parameters according to empirical methods and grid search, shown in table 1.

Table 1: parameters

| Optim | AdamW |
|---|---|
| weight_decay | 6e-5 |
| batch_size | 256 |
| lr | 3e-5 |
| Epoch | 3 |
| Top k in KNN | 100 |
| W1 | 0.6 |
| W2 | 0.4 |

- Evaluation metrics

We assess the similarity between the predicted embedding $E_{pred}$ and the ground truth embedding $E_{ground\ truth}$ by calculating their cosine similarity. The cosine similarity is defined as follows:

$$similarity = \frac{E_{pred} \cdot E_{ground\ truth}}{\|E_{pred}\| \cdot \|E_{ground\ truth}\|}$$

- Experiment result

To evaluate our experiment's performance, we do compared competitions. The higher metric is, the better our model will be. The experiment result is shown in table 2. Our method owns the highest metric 0.612 among these models, which is 0.013, 0.055, 0.011 higher than Clip, Clip + KNN（top 10） respectively.

Table 2: Experiment result

| Models | Cosine Similarity |
|---|---|
| Clip | 0.547 |
| Clip + KNN (top 10) | 0.601 |
| Clip + KNN (top 100) | 0.612 |

## V.  Conclusion

In our paper, we do feature engineering and using CLIP and KNN to predict prompts corresponding to imges. We introduce related work in section II and the model in section III. In section IV, our experiment is stated. Our method owns the highest metric 0.612 among these models, which is 0.013, 0.055, 0.011 higher than Clip, Clip + KNN（top 10） respectively.

## VI.  Reference

[1]  Qiao T, Zhang J, Xu D, et al. Mirrorgan: Learning text-to-image generation by redescription[C]//Proceedings of the IEEE/CVF Conference on Computer Vision and Pattern Recognition. 2019: 1505-1514.

[2]  Li B, Qi X, Lukasiewicz T, et al. Controllable text-to-image generation[J]. Advances in Neural Information Processing Systems, 2019, 32.

[3]  Ramesh A, Pavlov M, Goh G, et al. Zero-shot text-to-image generation[C]//International Conference on Machine Learning. PMLR, 2021: 8821-8831.

[4]  Rombach R, Blattmann A, Lorenz D, et al. High-resolution image synthesis with latent diffusion models[C]//Proceedings of the IEEE/CVF conference on computer vision and pattern recognition. 2022: 10684-10695.

[5]  Kingma D P, Welling M. Auto-encoding variational bayes[J]. arXiv preprint arXiv:1312.6114, 2013.

[6]  Ronneberger O, Fischer P, Brox T. U-net: Convolutional networks for biomedical image segmentation[C]//Medical Image Computing and Computer-Assisted Intervention–MICCAI 2015: 18th International Conference, Munich, Germany, October 5-9, 2015, Proceedings, Part III 18. Springer International Publishing, 2015: 234-241.

[7]  Radford A, Kim J W, Hallacy C, et al. Learning transferable visual models from natural language supervision[C]//International conference on machine learning. PMLR, 2021: 8748-8763.

[8]  Christoph Schuhmann, Richard Vencu, Romain Beaumont, Robert Kaczmarczyk, Clayton Mullis, Aarush Katta, Theo Coombes, Jenia Jitsev, Aran Komatsuzaki,Open Dataset of CLIP-Filtered 400 Million Image-Text Pairs, NeurIPS Workshop 2021

[9]  Byeon, Minwoo and Park, Beomhee and Kim, Haecheon and Lee, Sungjun and Baek, Woonhyuk and Kim, Saehoon, COYO-700M: Image-Text Pair Dataset, 2022

[9]  Lin T Y, Maire M, Belongie S, et al. Microsoft coco: Common objects in context[C]//Computer Vision–ECCV 2014: 13th European Conference, Zurich, Switzerland, September 6-12, 2014, Proceedings, Part V 13. Springer International Publishing, 2014: 740-755.

[10]  Desai K, Kaul G, Aysola Z, et al. RedCaps: Web-curated image-text data created by the people, for the people[J]. arXiv preprint arXiv:2111.11431, 2021.